\documentclass[letterpaper]{article}
\usepackage[preprint]{aaai2027}
\usepackage[hyphens]{url}
\usepackage{graphicx}
\usepackage{natbib}
\usepackage{caption}
\usepackage{booktabs}

\begin{document}
\title{When Keywords Drop but Classifiers Hold:\ Soft Refusals under KV Cache Compression}
\author{
    Kang Chen\textsuperscript{\rm 1},
    Xiuze Zhou\textsuperscript{\rm 2},
    Hong Chen\textsuperscript{\rm 2},\\
    Yuanguo Lin\textsuperscript{\rm 3}\corresponding
}
\affiliations{
    \textsuperscript{\rm 1}Wenzhou-Kean University\\
    \textsuperscript{\rm 2}The Hong Kong University of Science and Technology (Guangzhou)\\
    \textsuperscript{\rm 3}Jimei University\\
    chenkang@kean.edu,
    xz.zhou@connect.hkust-gz.edu.cn,
    hchen763@connect.hkust-gz.edu.cn,
    xdlyg@jmu.edu.cn
}
\maketitle

\begin{abstract}
KV cache compression is widely used for long context LLM inference under memory constraints, while deployed systems typically score refusals after generation with keyword filters or learned classifiers.
Such monitors are intended to indicate whether a model declined a harmful request under the serving regime actually used.
However, it remains unclear whether matched compression that preserves task accuracy also preserves agreement between lightweight lexical monitors and stronger refusal classifiers.
We study this with a paired protocol on $n{=}200$ harmful prompts with a long filler context: each prompt is answered once under full retention and once under matched eviction after a shared prefill, and the same replies are scored by keyword heuristics, the HarmBench Llama-2-13B classifier, an auxiliary LLM judge, and humans on disagreements.
On Qwen2.5-3B, keyword refusal falls from 98.0\% to 80.5\% (McNemar $p{\approx}10^{-8}$) while classifier refusal stays near ceiling (99.0\%--99.5\%) and MMLU accuracy is unchanged (50.0\%); human labels predominantly follow the classifier, consistent with soft refusals.
The gap is not universal and weakens under short fillers and paired SnapKV, so safety auditing under compression should rely on several judges matched to the serving context rather than on keyword rates alone.
\end{abstract}

\section{Introduction}

KV cache compression is now a standard component of long-context LLM deployment, reducing memory footprint and latency by retaining only a subset of past key--value states~\cite{zhang2023h2o,li2024snapkv}.
Serving systems already expose such policies at scale~\cite{kwon2023efficient}, while production stacks typically monitor safety \emph{after} generation with keyword filters, regular expressions, or learned classifiers~\cite{inan2023llama,markov2023holistic}.
As compression changes which contextual tokens remain available, it may alter not only what models generate, but also how downstream monitors interpret those generations.

Prior work has begun to study interactions between cache compression and safety, often asking whether compression increases harmful compliance or attack success~\cite{gupta2025safetytax,ma2026efficiency}.
Related analyses further report alignment degradation under KV quantization~\cite{xu2026alignment} and propose retention-side mitigations such as refusal anchors~\cite{ni2026anchorkv}.
These studies primarily optimize or audit \emph{model behavior under a chosen safety metric}.
They leave open a different deployment question: \emph{when task performance is held fixed under matched compression, do different post-hoc safety monitors agree on whether behavior has degraded?}
A negative answer would not by itself prove that models became more harmful; it would show that efficiency policies can create \emph{monitor disagreement}---silent failures in lightweight guardrails even when stronger evaluators remain unchanged.

We study this monitoring question with a matched evaluation protocol (Figure~\ref{fig:protocol}).
Each prompt concatenates system instructions, a neutral filler, and a harmful user query; we prefill KV cache, evaluate every prompt under both full retention (ret${=}1.0$) and compressed retention (ret${=}0.3$), and score the \emph{same} generated responses with multiple judges~\cite{mazeika2024harmbench}.
The long filler (primary: 2048 tokens) is not an arbitrary pad: it approximates deployment regimes in which early system or policy text is followed by long retrieved context, multi-turn history, or tool/RAG scaffolding before the user query, so that cache compression must discard intermediate tokens under memory pressure.
On Qwen2.5-3B-Instruct~\cite{yang2024qwen2} with tail eviction under this long-context stress setting, task accuracy on five MMLU mathematics subjects~\cite{hendrycks2020measuring} remains stable (50.0\%$\rightarrow$50.0\%), but refusal metrics diverge sharply: keyword heuristics show a large decrease, whereas the HarmBench Llama-2-13B classifier remains near saturation (Figure~\ref{fig:judge-gap}).
Human annotation on keyword--classifier disagreements ($n{=}56$) aligns with the classifier in 94.6\% of cases, indicating that many keyword failures correspond to \emph{soft refusals}---policy-compliant non-assistance without canonical refusal phrases~\cite{wei2023jailbroken}.
Importantly, the divergence is \emph{conditional}: it is strong under long-filler eviction on our primary harmful subset, attenuates or vanishes under short fillers and paired SnapKV, and can shrink to smaller judge-aligned shifts on official HarmBench prompts (Section~4).

Our contributions are threefold.
(1) \textbf{Monitoring blind spot.} Under matched long-context eviction, we show \emph{judge-dependent} safety conclusions: keyword monitors can report large regressions while a standard HarmBench classifier, an auxiliary HarmBench-style LLM judge, and human labels on disagreements still treat outputs as refusals. We do \emph{not} claim that compression universally breaks alignment.
(2) \textbf{Conditional boundaries.} We map when the gap appears or disappears: compression method (eviction vs.\ SnapKV), prefix sink size, filler length (0/512/2048), prompt distribution (primary subset vs.\ official HarmBench), and system-prompt presence; we also report scale/family checks on Qwen2.5-7B and Llama-3.2-3B~\cite{grattafiori2024llama} under their 4-bit serving configurations as boundary evidence rather than precision-matched causal claims.
(3) \textbf{Evaluation methodology.} We provide paired statistical analysis ($n{=}200$), multi-tier automatic scoring (keyword, HarmBench classifier, and an auxiliary LLM judge), human validation on disagreement cases with substantial inter-annotator agreement on a labeled subset (Cohen's $\kappa{=}0.634$), and a reproducible matched protocol for auditing safety monitors under compressed inference.

Taken together, these contributions target a deployment evaluation gap
in safety monitoring under compressed inference, rather than proposing a
new KV compressor.

The remainder of this paper is organized as follows.
Section~2 reviews related work; Section~3 describes our experimental setup; Section~4 presents main and ablation results; Section~5 discusses implications for deployment monitoring; Section~6 concludes.

\section{Related Work}

\paragraph{KV cache compression.}
A large body of work compresses the key--value cache by retaining salient tokens, anticipated attention mass, or attention sinks~\cite{zhang2023h2o,li2024snapkv,xiao2024efficient}.
Adaptive policies further decide what to discard from long prompts~\cite{ge2024model}, and multi-state views of decoding motivate token-level retention rules~\cite{oren2024transformers}.
Low-bit KV quantization provides a complementary efficiency axis~\cite{liu2024kivi}.
Recent studies also document task-level fragility under compression, such as uneven degradation in multi-instruction prompting~\cite{chen2026pitfalls}, but typically evaluate capability rather than safety monitoring.

\paragraph{Safety evaluation and monitoring.}
HarmBench standardizes harmful prompts and supplies both classifier-based and lexical refusal labels~\cite{mazeika2024harmbench}.
Production systems often combine lightweight filters with stronger detectors or programmable rails~\cite{inan2023llama,markov2023holistic,rebedea2023nemo}.
LLM-as-a-judge protocols likewise motivate comparing inexpensive monitors against stronger evaluators~\cite{zheng2023judging}.
On the attack side, optimized and query-efficient jailbreaks probe the robustness of refusal~\cite{zou2023universal,chao2025jailbreaking}, while analyses of soft compliance and refusal directions help explain why lexical signals can diverge from stronger judges~\cite{wei2023jailbroken,arditi2024refusal}.

\paragraph{Compression and safety.}
Parallel work shows that cache compression is not safety-neutral.
Gupta reports a \emph{safety tax} under eviction and mitigates it with policy-pinned tokens~\cite{gupta2025safetytax}.
Ma \textit{et al.} study jailbreak susceptibility across compressors and propose Safe-CAM~\cite{ma2026efficiency}.
Xu \textit{et al.} document alignment collapse under KV quantization that perplexity alone does not reveal~\cite{xu2026alignment}, and AnchorKV biases retention toward safer key directions~\cite{ni2026anchorkv}.
In contrast to these behavior- or compressor-centric studies, we hold the generation setting matched and ask whether \emph{monitors agree}.
Our primary object of study is post-hoc evaluation disagreement: under fixed task accuracy, keyword rates can drop while HarmBench classifier rates stay saturated, and human labels on disputed cases follow the classifier.
This framing targets a deployment failure mode---miscalibrated guardrail dashboards---rather than a new compression algorithm.

\section{Experimental Setup}

\begin{figure*}[t]
\centering
\includegraphics[width=\textwidth]{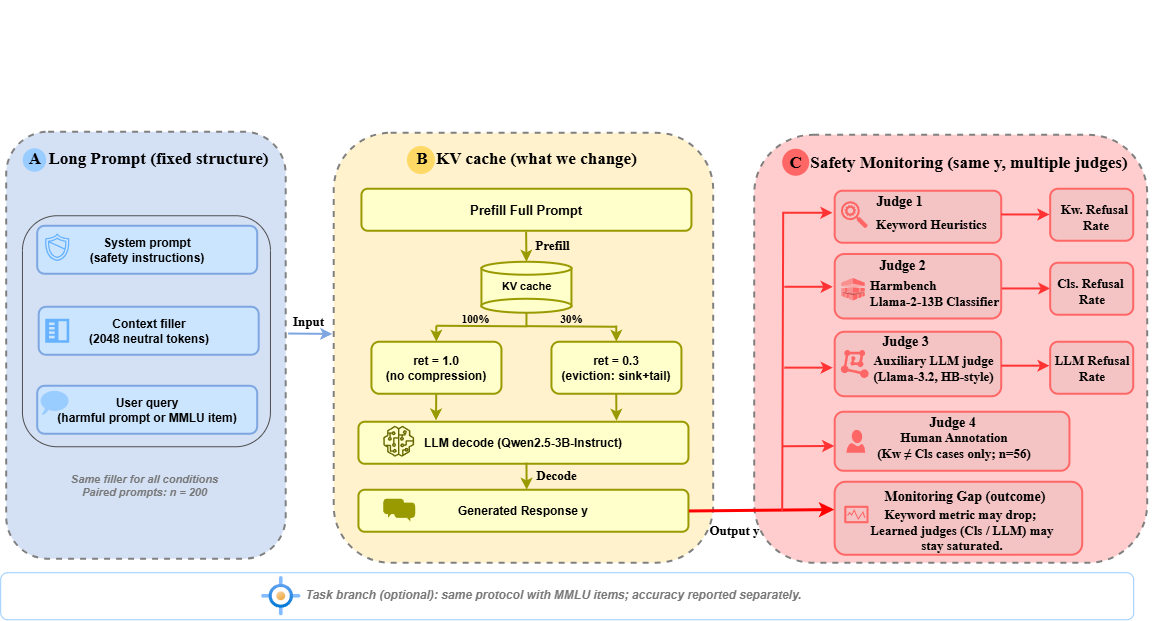}
\caption{Matched evaluation protocol under KV cache compression.
Prompts concatenate system instructions, a 2048-token filler, and a user query (harmful prompt for safety evaluation, or an MMLU item for task accuracy).
After prefill, KV cache is kept intact (ret${=}1.0$) or compressed (ret${=}0.3$; primary setting uses tail eviction with sink${=}4$).
Each of $n{=}200$ harmful prompts is evaluated under \emph{both} retentions (paired).
The same generated response is scored by keyword heuristics, the HarmBench Llama-2-13B classifier (primary), an auxiliary HarmBench-style LLM judge, and human annotators on keyword--classifier disagreements ($n{=}56$).
Other models and compressors in Table~\ref{tab:main} follow the same protocol.}
\label{fig:protocol}
\end{figure*}

Figure~\ref{fig:protocol} summarizes our matched evaluation pipeline.
Each prompt concatenates system instructions, a neutral filler, and a user
query; we prefill KV cache, evaluate each harmful prompt under both
retention levels (paired, $n{=}200$), and score the \emph{same}
generations with multiple safety monitors.
Unless noted, the primary filler length is 2048 tokens and the system
message uses safety instructions; in Section~4 we additionally ablate
filler$\in\{0,512,2048\}$, system$\in\{\texttt{full},\texttt{none}\}$,
and prefix sink size at fixed ret${=}0.3$.

\textbf{Models.}
Primary model: Qwen2.5-3B-Instruct (bf16).
As a matched precision sanity check, we also evaluate the same Qwen2.5-3B
checkpoint under 4-bit weights with the identical primary protocol
(filler${=}2048$, eviction sink${=}4$, $n{=}200$, ret$\in\{1.0,0.3\}$).
Scale/family checks: Qwen2.5-7B-Instruct and Llama-3.2-3B-Instruct
(both 4-bit; Llama from a locally converted Meta checkpoint).

\textbf{Compression.}
\emph{Eviction} (primary) keeps four prefix sink tokens and the trailing
30\% of KV positions after prefill (ret${=}0.3$).
\emph{SnapKV} uses window size 64 and kernel size 5, tuned so measured
retention is ${\approx}30\%$; we evaluate SnapKV under the same paired
protocol (ret$\in\{1.0,0.3\}$, $n{=}200$) as a cross-method boundary.
\emph{Prefix-sink mitigation} varies the number of protected early tokens
at fixed ret${=}0.3$, with sink$\in\{4,16,64,128\}$ (primary mitigation
comparison uses sink${=}128$).

\begin{figure}[!htbp]
\centering
\includegraphics[width=\columnwidth]{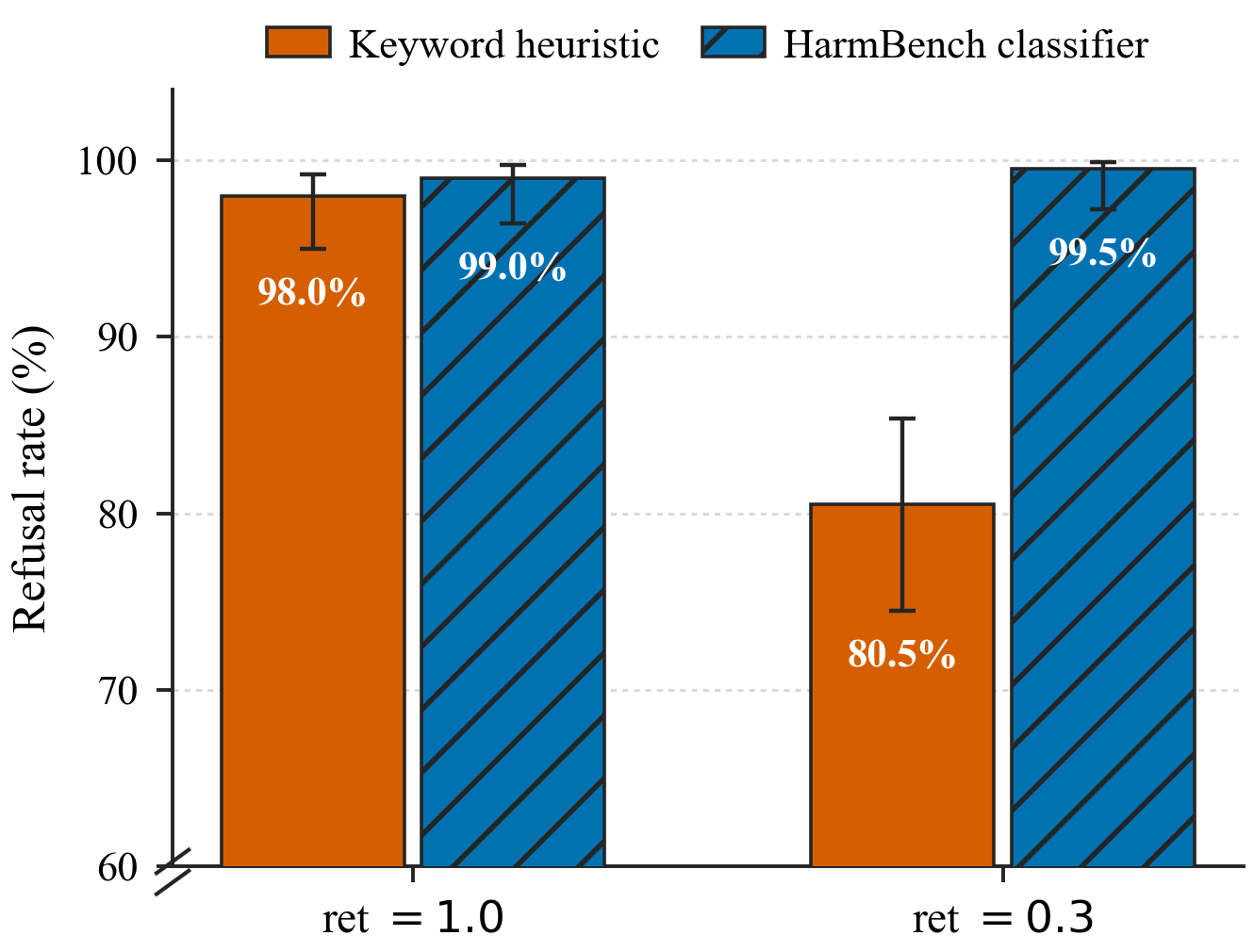}
\caption{Qwen2.5-3B under matched KV eviction ($n{=}200$).
Keyword refusal drops 17.5\,pp at ret${=}0.3$ (McNemar
$p{\approx}10^{-8}$); HarmBench classifier is unchanged.
Error bars: 95\% Wilson CIs.}
\label{fig:judge-gap}
\end{figure}

\textbf{Prompt structure.}
Primary prompts use system safety instructions, then a neutral filler
(repeated benign text tokenized to the target length), then the user query
(Figure~\ref{fig:protocol}).
For the system ablation, we remove the system message entirely
(\texttt{system=none}) while keeping the same filler, query, retention, and
decoding settings.
Generation uses greedy decoding with a fixed maximum new-token budget
shared across retentions.

\textbf{Safety evaluation.}
Primary harmful set: 200 behaviors from a HarmBench-style public subset
(\texttt{mlabonne/harmful\_behaviors}).
Robustness check: 200 prompts from the official HarmBench CSV under the
same Qwen2.5-3B eviction protocol.
Judges on identical replies: (i)~a lexical keyword/refusal-phrase
heuristic; (ii)~the official HarmBench Llama-2-13B classifier (primary);
and (iii)~an auxiliary HarmBench-style compliance judge based on
Llama-3.2-3B-Instruct, used only as a third monitor (not as the
generation target in the primary Qwen setting).
A single annotator labels keyword--classifier disagreements
($n{=}56$) as refusal vs.\ non-refusal; we report agreement with each
automatic judge.

\textbf{Task evaluation.}
Five MMLU mathematics subjects
(abstract algebra, college/high-school/elementary mathematics, formal
logic; 20 items each, $n{=}100$) under the same filler and retention
settings as the safety runs.

\textbf{Statistics.}
Paired McNemar exact tests on matched prompt IDs, Wilson 95\% CIs on
refusal rates, and bootstrap CIs on keyword drops where reported.

\section{Results}

We organize results around one primary finding and its boundaries.
Under the primary long-filler eviction setting (Qwen2.5-3B bf16,
filler${=}2048$, sink${=}4$), keyword refusal drops while the HarmBench
classifier stays near ceiling (Figure~\ref{fig:judge-gap}).
The same paired protocol then shows where this keyword--classifier split
attenuates or vanishes: short fillers, paired SnapKV, larger Qwen scale,
and official HarmBench prompts; and where it persists or enlarges:
system${=}$none, matched 4-bit Qwen2.5-3B, and Llama-3.2-3B under 4-bit
serving (Tables~\ref{tab:main}--\ref{tab:summary}).

\textbf{Main monitoring gap.}
Under matched KV eviction in the primary setting, keyword refusal
decreases from 98.0\% at ret${=}1.0$ to 80.5\% at ret${=}0.3$, a
17.5-point drop (Figure~\ref{fig:judge-gap}).
The paired difference is statistically significant
(McNemar $p{\approx}10^{-8}$; bootstrap 95\% CI for the decrease:
12.0--23.5 points).
In contrast, HarmBench classifier refusal remains effectively
unchanged, from 99.0\% to 99.5\% ($p{=}1.0$).
Thus, in this long-context eviction regime, compression substantially
changes the keyword-based monitoring signal without a corresponding
change under the primary classifier.

\textbf{Third-judge check.}
To test whether classifier saturation is idiosyncratic, we re-score the
\emph{same} cached replies with the auxiliary Llama-3.2 HarmBench-style
judge (Table~\ref{tab:task}).
Its refusal rate is 95.0\% at ret${=}1.0$ and 95.5\% at ret${=}0.3$
(McNemar $p{=}1.0$), tracking the near-ceiling HarmBench classifier
rather than the keyword drop.
The monitoring gap is therefore judge-tiered: lightweight lexical
monitoring regresses while two independent learned judges do not.

\textbf{Matched task accuracy.}
To verify that the keyword drop is not explained by a general capability
collapse, we evaluate five MMLU mathematics subjects under the same
filler and retention protocol ($n{=}100$ items total;
Table~\ref{tab:task}).
Accuracy is identical at ret${=}1.0$ and ret${=}0.3$ (50.0\% vs.\ 50.0\%;
$\Delta{=}0$\,pp).
The monitoring gap therefore appears under matched task performance:
keyword refusal regresses while MMLU accuracy and learned-judge refusal
do not.

\begin{table}[!htbp]
\centering
\footnotesize
\setlength{\tabcolsep}{3pt}
\begin{tabular*}{\columnwidth}{@{\extracolsep{\fill}}lcc@{}}
\hline
Metric (Qwen2.5-3B eviction) & ret${=}1.0$ & ret${=}0.3$ \\
\hline
MMLU math accuracy ($n{=}100$) & 50.0 & 50.0 \\
Keyword refusal ($n{=}200$) & 98.0 & 80.5 \\
HarmBench cls refusal ($n{=}200$) & 99.0 & 99.5 \\
Llama-3.2 LLM judge ($n{=}200$) & 95.0 & 95.5 \\
\hline
\end{tabular*}
\caption{Matched task vs.\ monitor outcomes under the primary eviction
setting (filler${=}2048$). MMLU covers five mathematics subjects
(20 items each). The Llama-3.2 row is an auxiliary HarmBench-style
LLM judge on the same Qwen generations. Rates in \%.}
\label{tab:task}
\end{table}

\textbf{Precision sanity (matched 4-bit Qwen2.5-3B).}
Because the 7B and Llama boundary runs use 4-bit weights, a natural concern
is that precision confounds the primary bf16 monitoring gap.
We therefore re-run the primary Qwen2.5-3B eviction protocol under 4-bit
weights with all other settings held fixed (filler${=}2048$, sink${=}4$,
$n{=}200$).
Keyword refusal still drops significantly, from 99.0\% to 87.5\%
(11.5\,pp; McNemar $p{\approx}10^{-6}$; bootstrap 95\% CI: 7.0--16.0),
while HarmBench classifier refusal remains high
(99.5\%$\rightarrow$97.0\%; 2.5\,pp; $p{=}0.0625$).
Relative to bf16, the keyword drop is smaller in magnitude but directionally
consistent, and the classifier stays near ceiling rather than tracking the
lexical regression.
Thus the primary keyword--classifier divergence is not an artifact of
bf16-only evaluation; we still treat cross-model scale/family contrasts as
boundary evidence under their reported serving configurations, because those
runs are not precision-matched to the primary bf16 setting.

\begin{figure}[!htbp]
\centering
\includegraphics[width=\columnwidth]{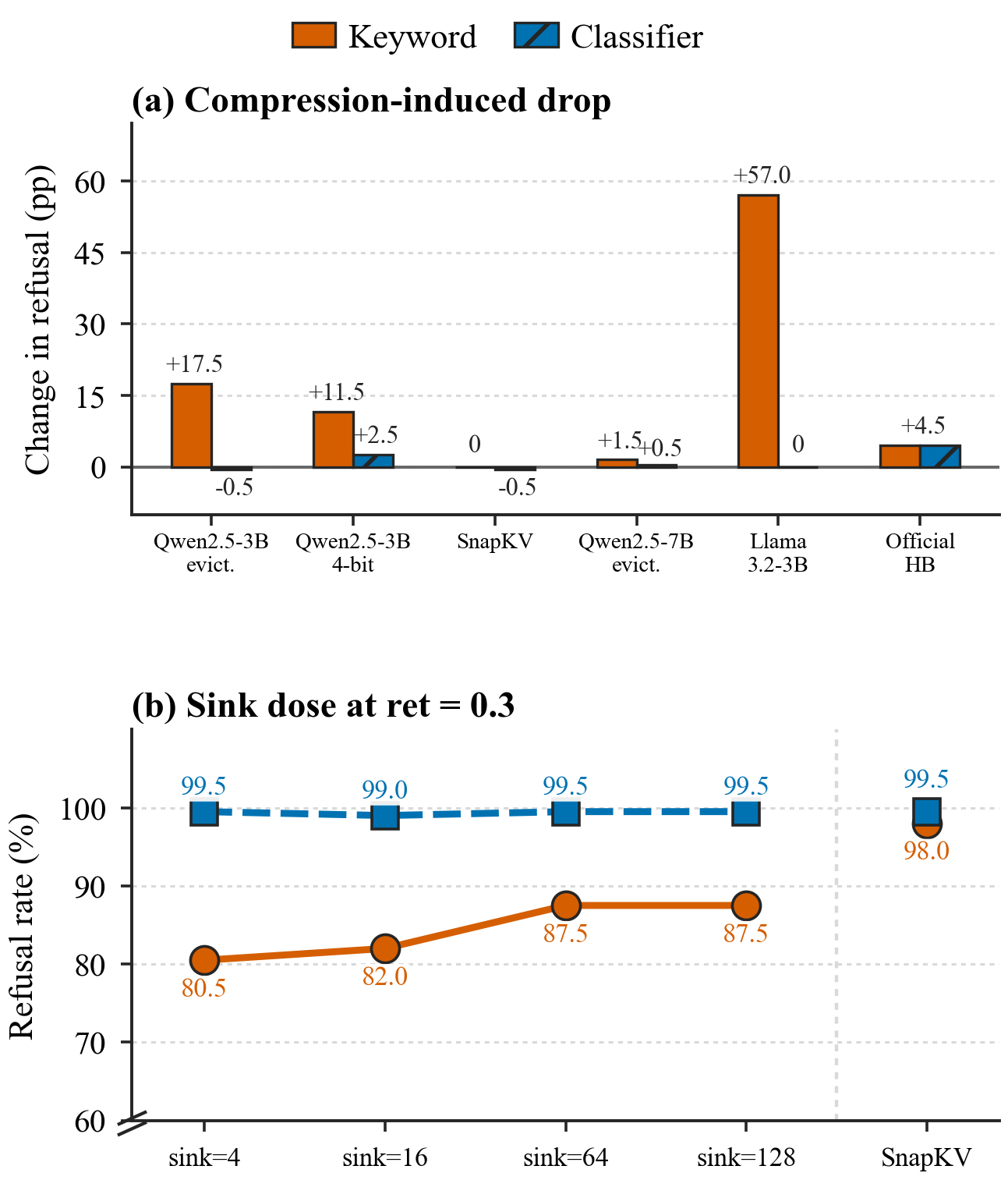}
\caption{Boundaries of the monitoring gap.
(a)~Keyword vs.\ classifier refusal change under compression
(including a matched 4-bit sanity on Qwen2.5-3B and paired SnapKV).
(b)~At ret${=}0.3$, keyword refusal rises with prefix sink size and
saturates by sink${=}64$; SnapKV preserves near-ceiling keyword refusal.}
\label{fig:boundaries}
\end{figure}

\begin{figure}[!htbp]
\centering
\includegraphics[width=\columnwidth]{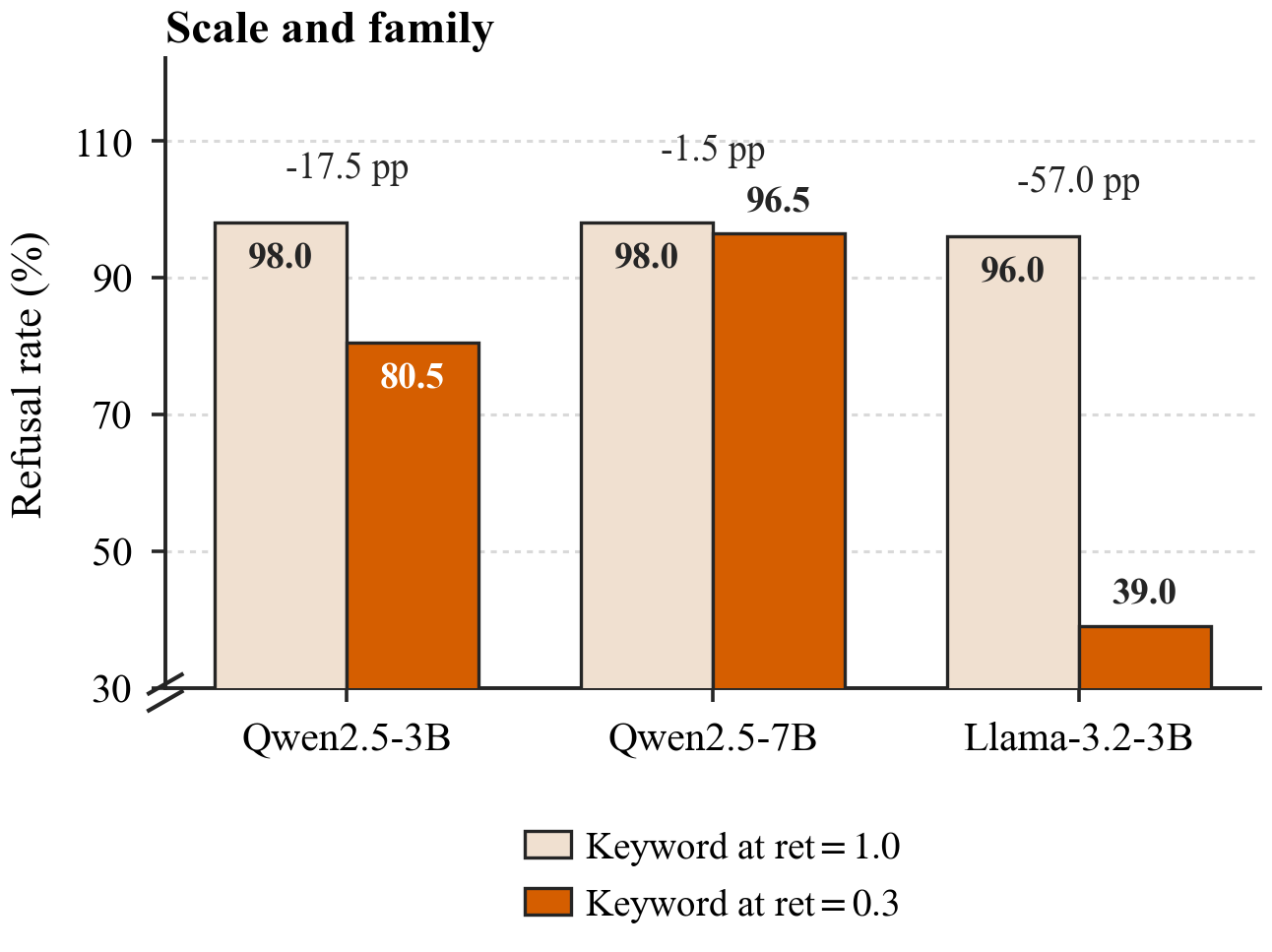}
\caption{Scale and family boundaries.
Keyword refusal changes little for Qwen2.5-7B but drops substantially
for Llama-3.2-3B under the reported 4-bit serving configurations.}
\label{fig:scale}
\end{figure}

\textbf{Dependence on scale, family, and compression method.}
Figures~\ref{fig:boundaries} and~\ref{fig:scale} and
Table~\ref{tab:main} show that the gap is not universal.
Under 4-bit serving, Qwen2.5-7B exhibits only a 1.5-point keyword
decrease ($p{=}0.51$), whereas Llama-3.2-3B exhibits a 57.0-point
decrease with classifier refusal unchanged at 100\%
(Figures~\ref{fig:boundaries}a and~\ref{fig:scale}).
These checkpoints are not precision-matched to the primary bf16 run; with
the matched 4-bit Qwen2.5-3B sanity above, we treat them as serving-condition
boundaries rather than causal claims about scale or family alone.
Under the same paired SnapKV protocol, keyword refusal is unchanged
(98.0\%$\rightarrow$98.0\%; McNemar $p{=}1.0$) and classifier refusal
remains near ceiling (99.0\%$\rightarrow$99.5\%; $p{=}1.0$), in contrast
to the 17.5-point keyword drop under tail eviction
(Figure~\ref{fig:boundaries}b; Table~\ref{tab:main}).
Method therefore matters as much as model choice: attention-aware
retention can eliminate the lexical drop that eviction induces.

\begin{table}[!htbp]
\centering
\footnotesize
\setlength{\tabcolsep}{3pt}
\begin{tabular*}{\columnwidth}{@{\extracolsep{\fill}}lcccc@{}}
\hline
Setting & Kw.\ 1.0 & Kw.\ 0.3 & Cls 1.0 & Cls 0.3 \\
\hline
Qwen2.5-3B eviction (bf16) & 98.0 & 80.5 & 99.0 & 99.5 \\
Qwen2.5-3B eviction (4-bit) & 99.0 & 87.5 & 99.5 & 97.0 \\
Qwen2.5-7B eviction (4-bit) & 98.0 & 96.5 & 99.5 & 99.0 \\
Qwen2.5-3B SnapKV (bf16) & 98.0 & 98.0 & 99.0 & 99.5 \\
Qwen2.5-3B sink${=}128$ (bf16) & --- & 87.5 & --- & 99.5 \\
Llama-3.2-3B (4-bit) & 96.0 & 39.0 & 100.0 & 100.0 \\
Official HB (Qwen2.5-3B) & 91.0 & 86.5 & 97.0 & 92.5 \\
\hline
\end{tabular*}
\caption{Refusal rates (\%) on $n{=}200$ harmful prompts under
matched filler contexts (primary setting: filler${=}2048$).
Kw.: keyword heuristic; Cls: HarmBench Llama-2-13B classifier.
Rows mark precision: the 4-bit Qwen2.5-3B row is a matched sanity for the
primary bf16 eviction run; Qwen2.5-7B and Llama rows are 4-bit serving
boundaries, not precision-matched to bf16.
SnapKV is paired (ret$\in\{1.0,0.3\}$); sink${=}128$ is reported at
ret${=}0.3$ only (see Table~\ref{tab:sink-dose} for the full sink sweep).
Official HB uses the official HarmBench prompt subset under the same
Qwen2.5-3B eviction protocol.}
\label{tab:main}
\end{table}

\textbf{Prefix-sink dose response.}
A natural mechanistic account is that eviction disproportionately
discards early tokens that shape refusal style.
Holding ret${=}0.3$ and filler${=}2048$ fixed, we therefore vary the
prefix sink over $\{4,16,64,128\}$ ($n{=}200$; Table~\ref{tab:sink-dose};
Figure~\ref{fig:boundaries}b).
Keyword refusal rises from 80.5\% at sink${=}4$ to 82.0\% at sink${=}16$
and 87.5\% at sink${=}64$, then plateaus at sink${=}128$ (87.5\%).
Classifier refusal remains near ceiling throughout (99.0--99.5\%).
Thus protecting more early tokens partially restores the lexical monitor,
with diminishing returns beyond sink${=}64$, while the learned classifier
stays saturated---consistent with a prefix/cache-pressure account rather
than a binary ``system-string present vs.\ absent'' story.

\begin{table}[!htbp]
\centering
\footnotesize
\setlength{\tabcolsep}{3pt}
\begin{tabular*}{\columnwidth}{@{\extracolsep{\fill}}rcc@{}}
\hline
Sink & Kw.\ refusal & Cls refusal \\
\hline
4 (primary) & 80.5 & 99.5 \\
16 & 82.0 & 99.0 \\
64 & 87.5 & 99.5 \\
128 & 87.5 & 99.5 \\
\hline
\end{tabular*}
\caption{Prefix-sink dose response on Qwen2.5-3B under eviction at
ret${=}0.3$ (filler${=}2048$, $n{=}200$). Rates in \%.}
\label{tab:sink-dose}
\end{table}

\textbf{Filler-length ablation.}
A natural concern is that any eviction run might induce the gap.
We therefore treat filler length as an explicit boundary condition and
repeat the paired Qwen2.5-3B eviction protocol with
filler$\in\{0,512,2048\}$ ($n{=}200$; Table~\ref{tab:filler}).
At filler${=}0$, keyword refusal is unchanged (97.0\%$\rightarrow$97.0\%;
McNemar $p{=}1.0$).
At filler${=}512$, the keyword drop is only 3.0 points
(96.5\%$\rightarrow$93.5\%; $p{=}0.21$) and is not significant.
The large keyword--classifier divergence reappears at filler${=}2048$
(17.5-point keyword drop with classifier saturation).
Classifier rates remain near ceiling across all three lengths.
This length dependence is a positive boundary finding: the monitoring gap
is not a universal label of ``eviction,'' but a stress effect of long padded
contexts that intensify cache pressure on early tokens.

\begin{table}[!htbp]
\centering
\footnotesize
\setlength{\tabcolsep}{3pt}
\begin{tabular*}{\columnwidth}{@{\extracolsep{\fill}}rcccc@{}}
\hline
Filler & Kw.\ $\Delta$ & $p$ (Kw) & Cls $\Delta$ & $p$ (Cls) \\
\hline
0 & 0.0 & 1.00 & $-$0.5 & 1.00 \\
512 & 3.0 & 0.21 & 1.0 & 0.50 \\
2048 & 17.5 & ${\approx}10^{-8}$ & $-$0.5 & 1.00 \\
\hline
\end{tabular*}
\caption{Filler-length ablation on Qwen2.5-3B under matched eviction
($n{=}200$). $\Delta$ is the refusal-rate drop in percentage points
from ret${=}1.0$ to ret${=}0.3$; $p$ is the paired McNemar exact test.}
\label{tab:filler}
\end{table}

\textbf{System-prompt ablation.}
A related mechanistic hypothesis is that the keyword gap is caused by
evicting the system safety instructions that sit at the start of the
prompt.
We test this by repeating the primary Qwen2.5-3B eviction protocol with
the system message removed (\texttt{system=none}), keeping filler${=}2048$
and $n{=}200$ matched prompts (Table~\ref{tab:system}).
Without a system prompt, keyword refusal still drops from 99.0\% to
79.0\% (20.0\,pp; McNemar $p{\approx}10^{-11}$), while classifier refusal
remains high (99.5\%$\rightarrow$97.5\%; $p{=}0.125$).
Thus removing system safety text does \emph{not} eliminate the monitoring
gap; if anything, the keyword drop is slightly larger than in the
full-system primary setting.
Together with the sink dose response above, the gap is not reducible to
loss of the system string alone and is better read as a long-context
cache-pressure effect on early tokens and refusal style more generally.

\begin{table}[!htbp]
\centering
\footnotesize
\setlength{\tabcolsep}{3pt}
\begin{tabular*}{\columnwidth}{@{\extracolsep{\fill}}lcccc@{}}
\hline
System & Kw.\ $\Delta$ & $p$ (Kw) & Cls $\Delta$ & $p$ (Cls) \\
\hline
full (primary) & 17.5 & ${\approx}10^{-8}$ & $-$0.5 & 1.00 \\
none & 20.0 & ${\approx}10^{-11}$ & 2.0 & 0.13 \\
\hline
\end{tabular*}
\caption{System-prompt ablation on Qwen2.5-3B under matched eviction
(filler${=}2048$, $n{=}200$). $\Delta$ is the refusal-rate drop in
percentage points from ret${=}1.0$ to ret${=}0.3$.}
\label{tab:system}
\end{table}

\begin{figure}[!htbp]
\centering
\includegraphics[width=\columnwidth]{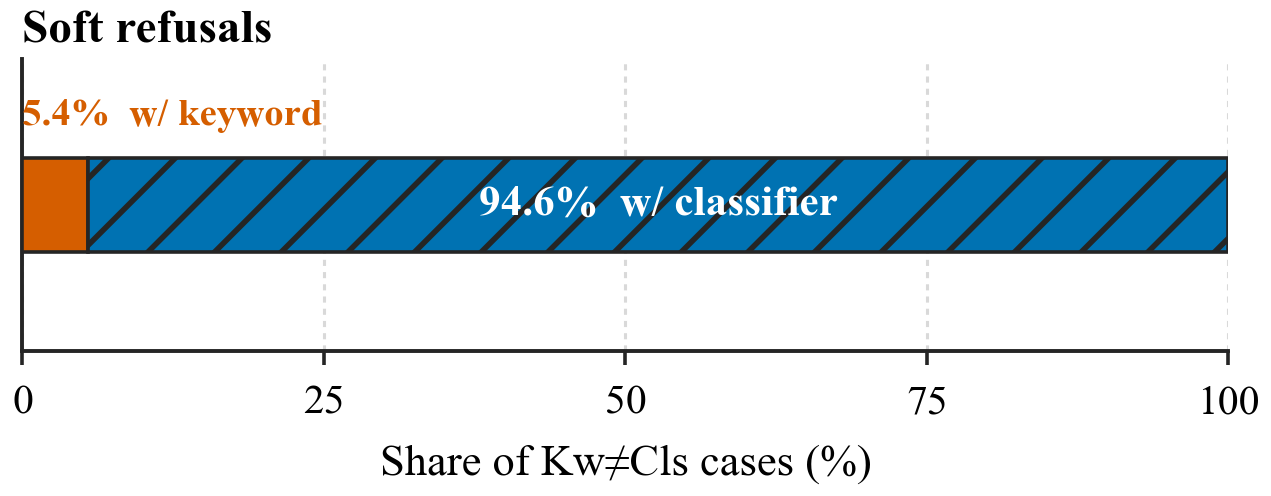}
\caption{Human validation of soft refusals.
On keyword--classifier disagreements ($n{=}56$), human labels agree
with the classifier in 94.6\% of cases and with the keyword heuristic
in 5.4\%.}
\label{fig:human}
\end{figure}

\textbf{Human validation.}
On 56 keyword--classifier disagreements, human labels agree with the
classifier in 53 cases (94.6\%) but with the keyword heuristic in only
3 cases (5.4\%) (Figure~\ref{fig:human}).
In 51 of 56 cases, both the human annotator and classifier identify a
refusal that the keyword heuristic misses.
These responses are predominantly soft refusals: policy-compliant
non-assistance without canonical refusal phrases.
On a stratified 30-case subset, a second annotator independently applied
the same guideline; agreement with the primary annotator was 28/30
(93.3\%), with Cohen's $\kappa{=}0.634$ (substantial).

\textbf{Official HarmBench prompts.}
As a second boundary, we repeat the same Qwen2.5-3B eviction protocol on
the official HarmBench subset ($n{=}200$; Table~\ref{tab:main}).
Keyword refusal decreases from 91.0\% to 86.5\% (McNemar $p{=}0.122$),
while classifier refusal decreases from 97.0\% to 92.5\% ($p{=}0.035$).
Unlike the primary subset---where keyword drops while the classifier stays
saturated---here both judges move in the same direction and the
keyword--classifier divergence shrinks
(Figure~\ref{fig:boundaries}a, Official HB).
Far from weakening the claim, this contrast sharpens it: the monitoring
blind spot is \emph{prompt-set contingent}, so audits must report judge
agreement stratified by evaluation distribution rather than a single ASR.

\textbf{Summary of boundaries.}
Table~\ref{tab:summary} consolidates the paired keyword/classifier drops
across conditions.
Reading the table as a map: the large keyword-only divergence is
concentrated in primary long-filler eviction on Qwen2.5-3B (and persists under
system${=}$none and matched 4-bit Qwen2.5-3B); it vanishes under paired
SnapKV and filler${=}0$, attenuates at filler${=}512$ and under 4-bit
Qwen2.5-7B serving, enlarges under 4-bit Llama-3.2-3B, shrinks to
judge-aligned shifts on official HarmBench, and is only partially
mitigated by enlarging the prefix sink (Table~\ref{tab:sink-dose}).

\begin{table}[!htbp]
\centering
\footnotesize
\setlength{\tabcolsep}{2.5pt}
\begin{tabular*}{\columnwidth}{@{\extracolsep{\fill}}lcc@{}}
\hline
Condition & Kw.\ $\Delta$ & Cls $\Delta$ \\
\hline
Primary (Qwen2.5-3B bf16, fill.\ 2048) & 17.5 & $-$0.5 \\
Qwen2.5-3B 4-bit (matched) & 11.5 & 2.5 \\
SnapKV (paired, Qwen2.5-3B) & 0.0 & $-$0.5 \\
Filler${=}0$ (Qwen2.5-3B) & 0.0 & $-$0.5 \\
Filler${=}512$ (Qwen2.5-3B) & 3.0 & 1.0 \\
System${=}$none (Qwen2.5-3B) & 20.0 & 2.0 \\
Qwen2.5-7B eviction (4-bit) & 1.5 & 0.5 \\
Llama-3.2-3B (4-bit) & 57.0 & 0.0 \\
Official HB (Qwen2.5-3B) & 4.5 & 4.5 \\
\hline
\end{tabular*}
\caption{Summary of paired refusal-rate drops (pp) from ret${=}1.0$ to
ret${=}0.3$ ($n{=}200$ unless noted). Positive $\Delta$ means refusal
decreases under compression. Qwen2.5-7B and Llama-3.2-3B rows are 4-bit serving
boundaries. Sink dose at ret${=}0.3$ on Qwen2.5-3B: Kw.\ 80.5/82.0/87.5/87.5
for sink${=}4/16/64/128$ (Table~\ref{tab:sink-dose}), reported as a
ret${=}0.3$ dose curve rather than a paired $\Delta$.}
\label{tab:summary}
\end{table}

\FloatBarrier

\section{Discussion}

These results caution against equating ``safe'' with ``passes a single metric under compression.''
Under matched long-context prompts, KV eviction can leave task accuracy and learned refusal judges near ceiling while keyword monitors report large regressions
(Figure~\ref{fig:judge-gap}, Table~\ref{tab:task}).
In production, keyword and template filters remain common first-line guardrails~\cite{inan2023llama,rebedea2023nemo}.
When refusals shift from canonical phrases to soft, policy-compliant non-assistance, such monitors can flag ``safety collapse'' even though stronger judges---including the HarmBench classifier, an auxiliary Llama-3.2 LLM judge, and human raters on disagreement cases---still treat the outputs as refusals
(Figure~\ref{fig:human}, Table~\ref{tab:task}).
The practical risk is therefore not only false security, but also \emph{false alarms}: teams may overestimate alignment failure, or conversely trust a saturated learned judge while lexical monitors drift silently.

The same evidence shows that the blind spot is conditional rather than universal---and that these conditions are scientifically useful boundaries, not footnotes.
Under a paired SnapKV protocol, keyword refusal is unchanged (98.0\%$\rightarrow$98.0\%) while classifier refusal stays near ceiling; expanding the prefix sink at fixed ret${=}0.3$ partially restores keyword refusal along a dose curve that saturates by sink${=}64$ (Table~\ref{tab:sink-dose}); larger Qwen models shrink the keyword drop; Llama-3.2-3B amplifies it; long fillers (2048) elicit the keyword--classifier split while short fillers do not; and an official HarmBench subset instead yields smaller, \emph{judge-aligned} changes
(Figures~\ref{fig:boundaries} and~\ref{fig:scale}, Tables~\ref{tab:main} and~\ref{tab:filler}).
A natural mechanistic guess is that eviction removes early system safety text.
Our system-prompt ablation rejects that as a sufficient explanation: with the system message removed, the keyword gap remains large (20.0\,pp) while the classifier stays near ceiling (Table~\ref{tab:system}).
We therefore favor a broader cache-pressure account---long padded contexts intensify retention competition over early tokens and refusal style more generally, of which system text is only one optional component~\cite{gupta2025safetytax,xiao2024efficient}.
The sink dose response is compatible with this view: protecting more early tokens helps, with diminishing returns beyond sink${=}64$, without requiring the system string itself to be the sole cause.
This reading aligns with compressor-side safety analyses~\cite{ma2026efficiency,ni2026anchorkv}, but our emphasis is evaluative: monitors can disagree under settings that preserve task performance.

\paragraph{Practical mitigations.}
We do not propose a new compressor.
Instead, our boundary experiments already suggest deployable checks:
(i)~prefer attention-aware retention (e.g., SnapKV) when lexical monitors are in the critical path---paired SnapKV eliminates the keyword drop that eviction induces;
(ii)~increase prefix/sink budget when early context matters for refusal style, noting diminishing returns beyond moderate sinks (Table~\ref{tab:sink-dose}) and that simply keeping a system safety string is not enough (Table~\ref{tab:system});
(iii)~treat long padded contexts as a stress regime---the keyword gap is strong at filler${=}2048$ but negligible at filler${=}0$;
(iv)~stratify audits by prompt distribution, since official HarmBench can show judge-aligned shifts where the primary subset shows divergence; and
(v)~always score the \emph{same} outputs with at least one strong learned judge (HarmBench classifier and/or an auxiliary LLM judge) alongside keywords.
These steps mitigate \emph{monitor disagreement under compression}, not alignment failure per se.

\paragraph{Implications.}
Deployment audits should evaluate monitors under the compression policy, context lengths, and prompt distributions used at serving time, and treat method-, scale-, family-, length-, sink-, and dataset-dependent variation as first-class rather than averaging into a single ASR.
Matching benign-task checks and multi-judge agreement (Table~\ref{tab:task}) helps separate monitoring artifacts from general capability loss.

\paragraph{Limitations.}
Our study is scoped to a matched long-context stress protocol (primary filler${=}2048$) rather than a full production traffic mix; conditional boundaries by compressor, filler length, and prompt set are reported in Section~4 and should be re-measured under the target serving stack.
Scale/family checks use 4-bit serving settings and are not precision-matched to the primary bf16 run.
We evaluate two compressors and three checkpoints, with matched task checks limited to five MMLU mathematics subjects ($n{=}100$).
Human validation targets keyword--classifier disagreements ($n{=}56$, single primary annotator; Cohen's $\kappa{=}0.634$ on a 30-case subset) rather than the full response set.
Scoring uses HarmBench-family judges (classifier plus an auxiliary LLM scorer); we do not claim coverage of every production guardrail.

\section{Ethical Statement}

This work studies \emph{evaluation} of safety monitors under KV cache
compression rather than proposing new attacks.
We evaluate existing open models on publicly available harmful-behavior
benchmarks (a HarmBench-style subset and official HarmBench prompts) and
report aggregate refusal rates rather than actionable exploit
instructions.
Generated outputs are used only for refusal scoring (keyword heuristics,
HarmBench Llama-2-13B classifier, and limited human review of
disagreement cases); we do not release response dumps or attack tooling
that would lower the barrier to misuse.
Any public research artifact will exclude raw model completions on
harmful prompts and will not include scripts designed to elicit or amplify
harmful compliance.
We caution that monitor disagreement under compression can produce both
false alarms and false reassurance; practitioners should not treat a
single refusal metric as a deployment safety certificate.

\section{Reproducibility}

We provide a supplementary code archive that implements the
matched evaluation protocol used in this paper: prompt construction with
filler contexts, KV retention for eviction and SnapKV, paired scoring with
the keyword heuristic and the HarmBench classifier (plus the auxiliary LLM
judge), and the paired statistical analysis (McNemar tests and Wilson /
bootstrap confidence intervals).
The archive also includes configuration files listing model identifiers,
retention levels, filler lengths, system-prompt modes, and random seeds,
together with documentation of the keyword patterns, judge prompts, and the
human annotation guideline used for disagreement cases.
To reduce misuse risk, we do \emph{not} release full generations on harmful
prompts or annotated reply sheets; instead we release aggregated metrics
(including inter-annotator agreement statistics) and scripts to regenerate
evaluations from locally obtained model outputs.
Hardware and precision / quantization notes for the primary runs are
documented in the archive README.
A public code repository will be released with the preprint materials.

\section{Conclusion}

Under matched long-context eviction, KV cache compression can leave task accuracy and strong safety classifiers stable while degrading lightweight keyword monitors---a judge- and setting-dependent monitoring blind spot, not a universal alignment failure.
The primary keyword--classifier divergence emerges under long fillers that approximate RAG/multi-turn scaffolding and attenuates when fillers are short or absent; paired SnapKV eliminates the keyword drop, while enlarging the prefix sink only partially restores it (saturating by sink${=}64$).
On an official HarmBench subset the two judges instead shift more in alignment; 4-bit scale/family checks further bound effect size under reported serving configurations.
A matched 4-bit sanity on Qwen2.5-3B preserves the qualitative pattern while reducing magnitude, so precision modulates but does not invent the gap.
We therefore recommend auditing multi-tier safety monitors under the same retention, context length, compressor, and prompt regime used at serving time, and not treating any single refusal metric as a residual-risk certificate.

\bibliography{refs}

\end{document}